\documentclass[letterpaper, 10 pt, conference]{ieeeconf}  
\IEEEoverridecommandlockouts                              
\usepackage{listings}               
\usepackage[pdftex]{graphicx}       
\usepackage{xcolor}                 
\usepackage{tabularx}

\lstdefinestyle{Prolog} {language=Prolog,
                         lineskip=-0.3ex,
                         fontadjust=true,
                         basicstyle={\footnotesize\nopagebreak[4]},
                         commentstyle=\footnotesize, 
                         keywordstyle=\footnotesize,
                         showstringspaces=false,
                         showspaces=false,
                         showtabs=false,
                         moredelim=**[is][\bf]{@}{@},
                         moredelim=**[is][\it]{~}{~}
                         }
\lstdefinestyle{Lisp}   {language=Lisp,
                         lineskip=-0.5ex,
                         fontadjust=true,
                         basicstyle={\footnotesize \nopagebreak[4]},
                         commentstyle=\footnotesize,
                         morekeywords={achieve,def-goal,perform,detect,def-top-level-plan,with-designators,a,an,
                            location,object,action,at-location,with-failure-handling,motion,target},
                         keywordstyle=\bf\footnotesize,
                         moredelim=**[is][\bf]{@}{@},
                         moredelim=**[is][\it]{~}{~},
                         moredelim=**[is][\ldots]{!}{!}
                         }

\graphicspath{{img/}}

\title{\LARGE \bf 
Towards Plan Transformations for Real-World Pick and Place Tasks
}

\author{Gayane Kazhoyan$^{1*}$, Arthur Niedzwiecki$^{1*}$ and Michael Beetz$^{1}$\\
        {\{kazhoyan, aniedz, beetz\}@cs.uni-bremen.de} 
        \thanks{$^{1}$ The authors are with the Institute for Artificial Intelligence, 
                 University of Bremen, Germany.}
        \thanks{$^{*}$ Both authors contributed equally to this manuscript.}}

\begin{document}

\maketitle
\thispagestyle{empty}
\pagestyle{empty}

\begin{abstract}
In this paper, we investigate the possibility of applying plan transformations to general manipulation plans in order to specialize them to the specific situation at hand. We present a framework for optimizing execution and achieving higher performance by autonomously transforming robot's behavior at runtime. We show that plans employed by robotic agents in real-world environments can be transformed, despite their control structures being very complex due to the specifics of acting in the real world. The evaluation is carried out on a plan of a PR2 robot performing pick and place tasks, to which we apply three example transformations, as well as on a large amount of experiments in a fast plan projection environment.
\end{abstract}

\section{INTRODUCTION}

For the robot control programs to scale to open tasks and environments, they have to be general and consider a huge variety of situations.
In many cases, being general prevents the system from being able to exploit the specifics of the situation at hand, which often results in inefficient behavior.
Ideally, the robot would apply the knowledge that certain complications will not occur and do not have to be accounted for in the particular situation and discover opportunities in the context of the current task to be more efficient and robust.
One way of achieving both generality and being able to exploit the specifics of the situation is to enable the robot to reprogram its general plan towards the specific task and environment. Reprogramming has been studied substantially in the area of automated programming \cite{automatic-programming, rewriting-strategies}, however, transformation of robot control programs is particularly challenging because plans designed for real-world applications contain concurrent-reactive behaviors and comprehensive failure handling strategies to deal with sensor noise, partial observability of the world and non-determinism of actions.
In this paper, we investigate the possibility of applying plan transformations to general manipulation plans employed by robotic agents in real-world environments.

\begin{figure}[htb]
\centering
\includegraphics[width=0.99\columnwidth]{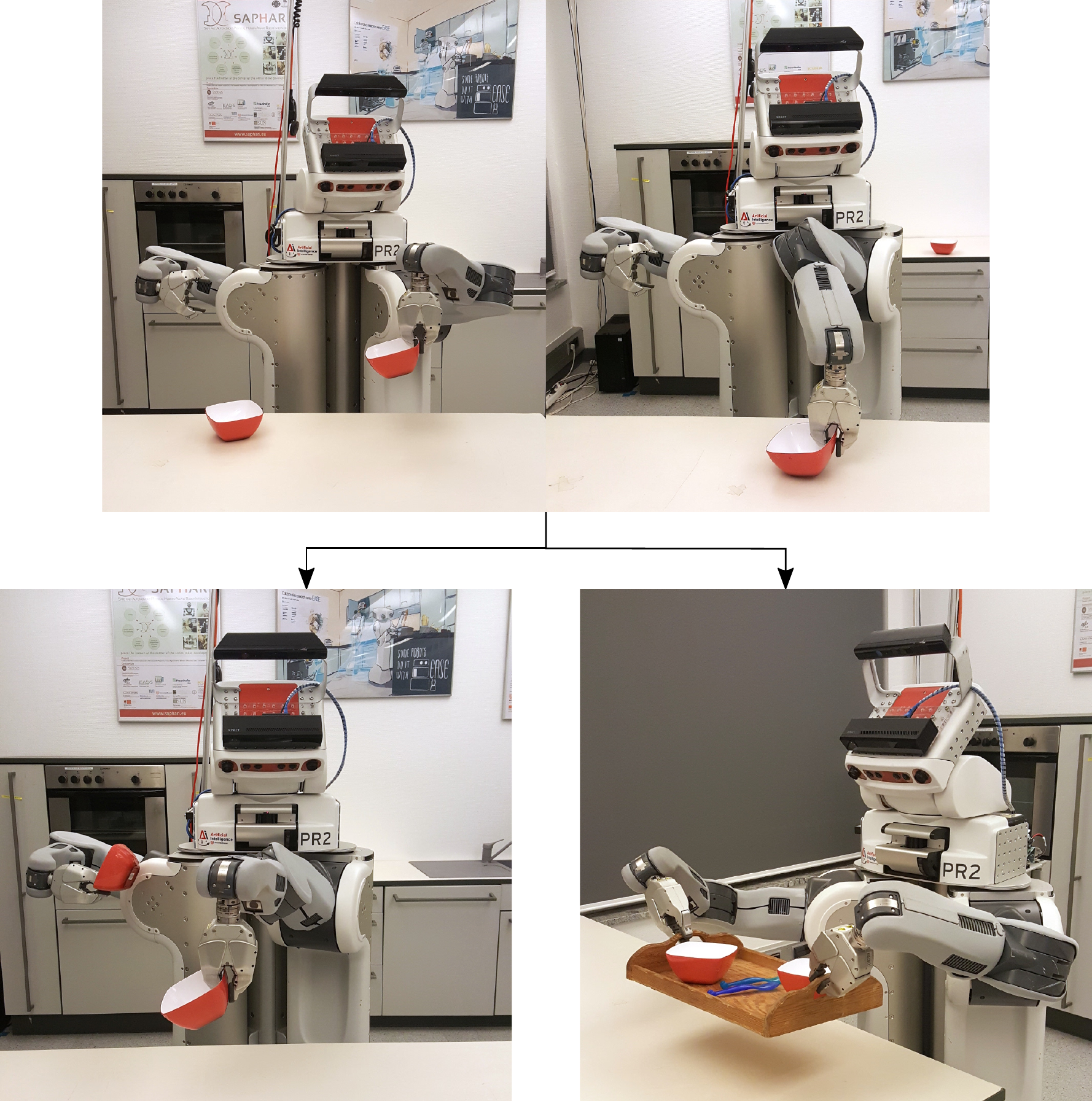}
\caption{(top) PR2 transporting two objects in a row with one hand each, \newline
         (bottom left) robot using both hands to transport objects, \newline
         (bottom right) robot utilizing a tray to transport multiple objects.}
\vspace{-2.1ex}
\label{fig:intro}
\end{figure}

Plan transformation is possibly the most powerful technique available for autonomously changing robot's course of action.
Presently, there exist very few well-designed robust manipulation plans that can handle broader varieties of tasks and contexts, executable in real-world settings.
Having a robot changing its plans automatically is even more difficult to imagine.
However, it is clear that there is a huge potential in the application of plan transformations: substantial changes in action strategies, such as transporting multiple objects with both hands or with a tray (see Figure \ref{fig:intro}), or leaving doors of shelves open while taking objects out, 
can be achieved by plan transformations without introducing unwanted side effects.

As an application domain we consider one of the most typical categories of mobile manipulation actions --- pick and place tasks --- although the framework is not limited to a specific task domain.
We present three example plan transformation rules that are applied to a plan consisting of tasks for setting a simple breakfast table and cleaning up. We also present a pipeline, which enables users to define new transformation rules and register them as plugins in the system, such that the new rules can automatically be applied to optimize robot's behavior during next execution.

The main contribution of this paper to the state of the art is the proof of concept and a working system that allows robotic agents acting in real-world environments to achieve performance gains by applying plan transformations to their general manipulation plans without introducing unwanted side effects.
The evaluation is carried out on a large amount of experiments in a fast plan projection environment to make statistically meaningful conclusions on improved execution performance, and on a real-world table setting and cleaning plan to demonstrate the feasibility of the approach.

\section{PLAN STRUCTURE AND TASK TREES}



The plans we use in our system are written in the CRAM Plan Language \cite{cram} for representing concurrent-reactive behavior, and we use the concept of \textit{entity descriptions} \cite{designators} for keeping plans general and modular. Listing \ref{lst:example-plan} shows a simple example plan for transporting two objects, one of type \textit{milk} and one of type \textit{cup}, using entity descriptions.

\begin{lstlisting}[style=Lisp, caption={Example plan for transporting two objects}, captionpos=b, label={lst:example-plan}]
(def-plan transport-objects ()
 (dolist (?type '(milk cup))
  (perform 
   (an action 
       (type transporting)
       (object (an object
                   (type ?type)))
       (location (a location
                    (on (an object
                            (type CounterTop)
                            (name sink_area)))))
       (target (a location
                  (on (an object
                          (type CounterTop)
                          (name island_area)))))))))
\end{lstlisting}
Through a loop, the variable \textit{?type} contains values \textit{milk} and then \textit{cup}. In the body of the loop, a \textit{transporting} action is performed for each of the objects, fetching the \textit{object} of type \textit{?type} from one \textit{location} and delivering it to the \textit{target} area.

A plan consists of a sequence (ordered, partially ordered or concurrent) of calls to perform actions, often secured by failure handling mechanisms for achieving reliable and stable behavior.
Performing an action within a plan is in its turn implemented through calling its respective subplan.
For example, \textit{transporting} an object from one location to another is implemented through the sequence of actions of searching for the object, fetching it and delivering, while fetching an object involves positioning the robot in reach of the object, then grasping and lifting it.
If the object cannot be found within the camera view of the robot or if it is unreachable from current location, failure recovery strategies command the robot to reposition itself, try grasping with another arm, etc. 
Thus, performing higher-level plans triggers execution of lower-level plans, which results in a hierarchy, eventually ending in atomic motions at the leaves of the hierarchy.
Motions resemble lowest-level atomic activities of the robot, such as moving an arm into a specific position or closing a gripper.
Entity descriptions are ground into robot's environment on demand. Thus, the programmer describes a location in the plan as an area, and during execution one pose is sampled from the area the moment the location entity is evaluated. If the chosen pose does not suffice for robot's current action, failure handling of the corresponding plan accounts for a different sample.
With these modular action descriptions and using the CRAM Plan Language we are able to construct complex scenarios of robot activities.

To analyze the robot's behavior and optimize it in the scope of any scenario and context, we make use of the \textit{task tree} datastructure, introduced in our previous work \cite{execution-trace}. The task tree is the runtime representation of the executed plan, containing all plan-relevant information, including specific parameters the plans were called with. 
At runtime, for each executed plan a node is automatically created in the task tree, containing references to its children nodes for every subplan executed within the plan. The top-level plan generates the root node of the tree, and the leaves are created when atomic motions are executed.

The task tree can be used as an extensive log for introspection of executed plans, e.g., to see which actions succeeded and which failed and what were the parameters used. However, task trees have, additionally, another important feature that we make use of for plan transformations, namely, the task tree persists between multiple executions of the same plan. Therefore, during the second execution of the plan, its task tree is not constructed anew but the old one is updated with the new plan arguments used in the current execution. Thus, transforming a task tree and re-executing its plan results in a transformed behavior on the robot. This means that the robot has to have the plan executed once before it can optimize the behavior through transformations. In our system, the robot first executes the plan in fast projection environment, then transforms the task tree and executes the same plan in the real world (see Section \ref{sec:experiments} for more details). As plan projection is much faster than realtime, this process does not significantly delay execution.

\begin{figure}[htb]
\centering
\includegraphics[width=\linewidth]{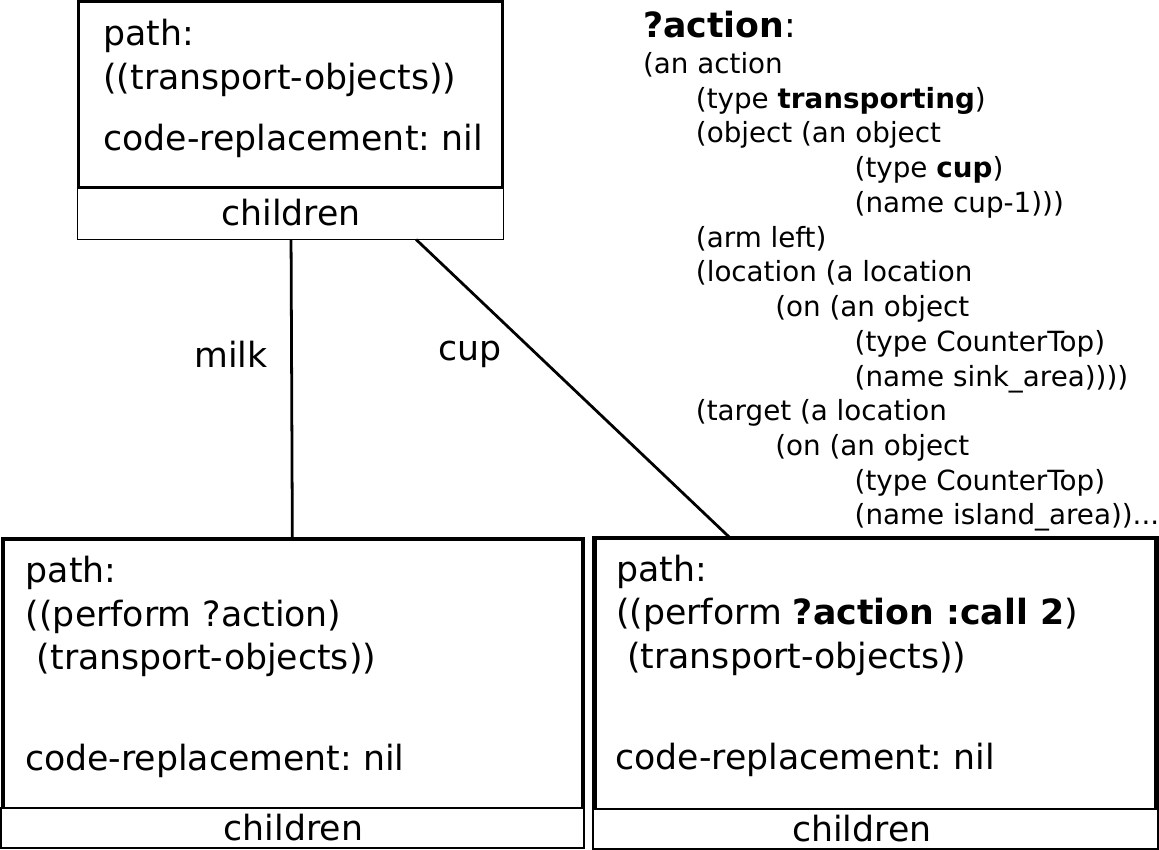}
\caption{Top two layers of a task tree of a plan for transporting two objects}
\label{fig:exampletasktree}
\vspace{-2ex}
\end{figure}

Figure \ref{fig:exampletasktree} shows the first two layers of the task tree generated from executing the plan from Listing \ref{lst:example-plan}.

The top-level node corresponds to the \textit{transport-objects} plan and has two child nodes.
Each node contains a path slot, which describes a unique path, with which a node can be found and referred to within the task tree.
It is built by concatenating the name of the node and all its parents. If there are multiple siblings with the same name (as in case of loops), \textit{:call N} is appended to make the path unique --- thus, \textit{:call 2} in the path of the second transporting action. All the parameters, with which the plan was called, are also stored in the task tree, thus, the \textit{?action} parameter of the second transporting action contains a \textit{cup} object, and the first one -- a \textit{milk}.

Every task node has a \textit{code-replacement} slot, which, together with paths and plan parameters, is used in the next section (see Section \ref{sec:transformations}) for transforming an executed plan.

\section{TRANSFORMATION RULES}
\label{sec:transformations}

\begin{figure}[htb]
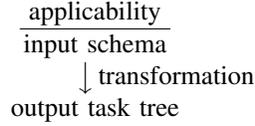

\centering
\begin{tabular}{c}
\underline{ applicability }\\
input schema \\
 $\left.%
 \begin{tabular}{r}
  \hspace{4em} \\
 \end{tabular}
 \right\downarrow \rotatebox[origin=c]{0}{transformation}$ \\
output task tree
\end{tabular}\caption{Workflow of applying a transformation rule.} \label{fig:muellertransform}
\vspace{-1ex}
\end{figure}

During plan transformation, the task tree of the plan at hand undergoes thorough analysis, whereby patterns in the behavior and opportunities for improvement are searched for in order to optimize robot's performance.
In our system, we follow the definition of plan transformations used in M\"uller's transformational framework \cite{muller2008transformational}. It is illustrated in Figure \ref{fig:muellertransform}.

\textit{Applicability} checks if a transformation is suitable for the current task tree, traversing the tree via Prolog predicates and searching for patterns that can be used in a transformation.
Applicability rules, additionally, extract information from the relevant nodes of the tree, such as node paths and relevant entity descriptions contained in the plan arguments, which is subsequently used by the transformation function. These data constitute the \textit{input schema} of the task tree.
The \textit{transformation} is a function that overwrites the \textit{code-replacement} slots of nodes in the task tree, by following the node paths and using the entity descriptions obtained from the input schema, resulting in an \textit{output task tree}.
When a plan is executed, the \textit{code-replacement} slot of the node corresponding to the currently active subplan is examined, and, when present, instead of the original code of the subplan the given code replacement is executed. 

By changing the content of single nodes instead of producing a completely new plan, we can modify the task tree in various ways without affecting the original plan, since the effect of applied transformations only manifests itself at runtime. On the other hand, the structure of the task tree cannot be altered, since it is based on the hierarchy of subplans in the original plan. This restriction poses advantages and disadvantages for the potential of plan transformations using task trees, which is discussed later in Section \ref{sec:conclusion}.

Below we describe three different transformation rules that have been implemented in the scope of our investigation to prove that plan transformations can be applied to real-world pick and place plans. The transformations differ from one another by the resources of the robot and the environment that they utilize for optimizing execution performance and by the effect they have on the task tree.

\subsection{Transporting Objects With Both Hands}

\begin{figure}[htb]
\centering
\includegraphics[width=0.7\linewidth]{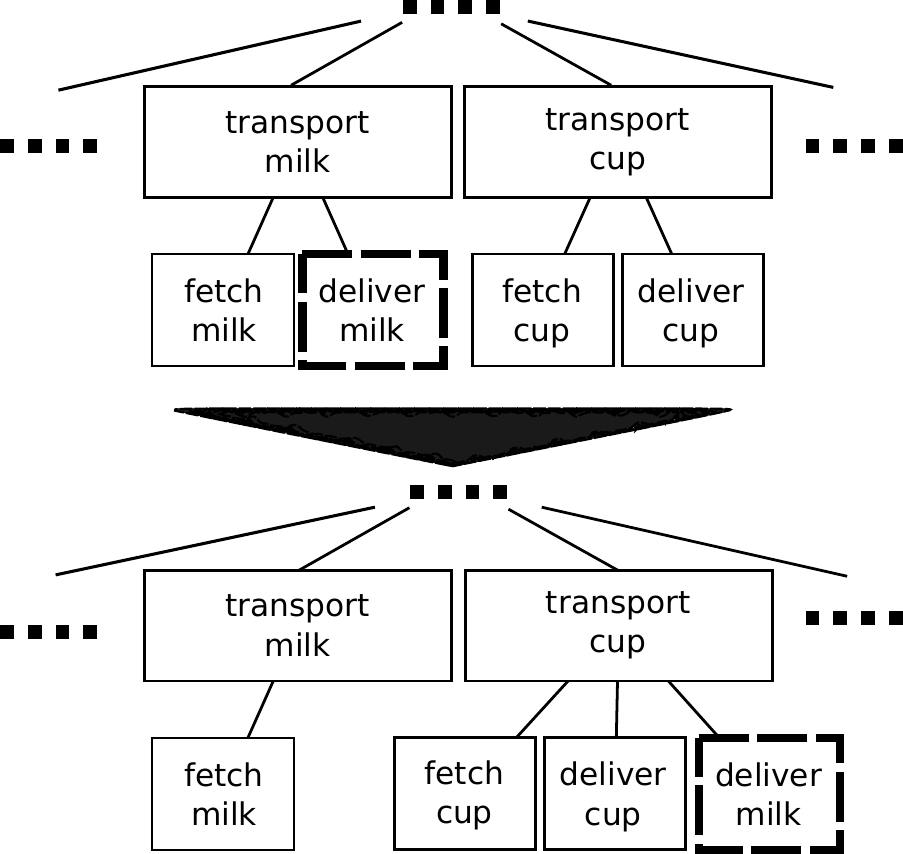}
\caption{Task tree for carrying two objects -- \textit{milk} and \textit{cup} -- transformed with \textit{both-hands-transform}. It delays the \textit{delivery} action of the first object \textit{milk} until the second object \textit{cup} has been delivered.}
\label{fig:scenariobhtransform}
\vspace{-1.5ex}
\end{figure}

The first transformation allows to optimize parts of a general plan that transport two objects in a row with one hand each time into transporting both objects at once. We call it \textit{both-hands-transform}.
It extracts two actions of type \textit{transporting} from the task tree that fetch from and deliver to areas described by the same location description, then switches the order of their \textit{fetching} and \textit{delivering} actions. Whereas in the original plan the robot first fetches one object and delivers it immediately, in the transformed plan the delivery of the first object is delayed and instead executed after the delivery of the second object (see Figure \ref{fig:scenariobhtransform}).

This transformation requires at least two transporting actions to be present in the plan. Fetching locations as well as delivering locations of both objects have to be within the same areas correspondingly, otherwise the robot could navigate a long distance, only to use his second hand.
The transporting actions in the original plan have to be performed by one arm.
Considering these restrictions, the applicability rule of \textit{both-hands-transform} is defined as following:

\begin{lstlisting}[style=Prolog]
both_hands_rule(Deliver_action, Deliver_path,
                Second_deliver_path) :-
  transport_task("top-level", First_task),
  transport_task("top-level", Second_task),
  not(First_task == Second_task),
  not_transformed(First_task),
  not_transformed(Second_task),
  task_action_param(First_task, First_action),
  task_action_param(Second_task, Second_action),
  one_arm_action(First_action),
  one_arm_action(Second_action),
  fetch_from_same_loc(First_action, Second_action),
  deliver_to_same_loc(First_action, Second_action),
  deliver_task(First_task, Deliver_task),
  task_action_param(Deliver_task, Deliver_action),
  task_path(Deliver_task, Deliver_path),
  deliver_task(Second_task, Second_deliver_task),
  task_path(Second_deliver_task,
            Second_deliver_path).
\end{lstlisting}

If the predicate finds suitable bindings, a matching pattern has been found and the input schema is returned as a triple: \textit{Deliver\_action} is the action description of delivering the first object, \textit{Deliver\_path} is the path to the corresponding node of the delivering action, and \textit{Second\_deliver\_path} points to the delivering action of the second object.
The predicate \textit{not\_transformed} ensures that tasks that have already been transformed, i.e., have their \textit{code-replacement} set, are not being transformed a second time. Optimizing task tree nodes that have already been transformed is rather challenging using code replacements, which is a limitation of our approach (see Section \ref{sec:conclusion} for related discussion).

Using the input schema, the transformation function finds the first delivering action under \textit{Deliver\_path}, replaces its code with \textit{nil} and appends code for performing \textit{Deliver\_action} at the end of the task at \textit{Second\_deliver\_path}.
 

\subsection{Avoiding Redundant Environment Manipulation}

\begin{figure}[htb]
\centering
\includegraphics[width=0.8\linewidth]{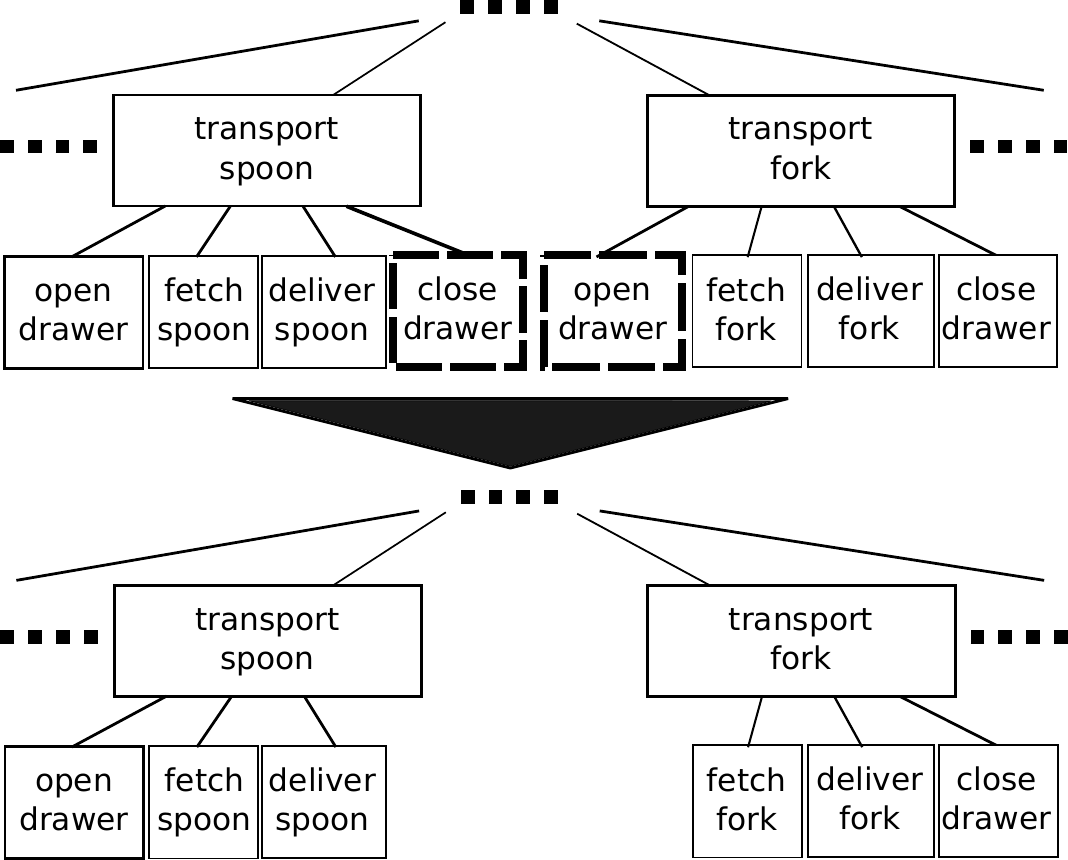}
\caption{Task tree for fetching cutlery from a drawer transformed with \textit{container-transform}. It removes all the opening and closing actions of the same container between the first \textit{open} and the last \textit{close}.}
\label{fig:scenario2environmenttransform}
\vspace{-1.5ex}
\end{figure}

This transformation, called \textit{container-transform}, optimizes a general plan of fetching objects out of drawers, cupboards and other containers, by skipping redundant opening and closing actions.
%
%
It can be applied to plans that include multiple \textit{transporting} actions, whose \textit{fetching} location is described using the \textit{in} keyword, i.e. actions that fetch an object from a container. For example, \textit{(a location (in (an object (type Drawer))))} describes a location in a drawer.
%
%
%
When applying the \textit{container-transform}, certain actions are simply discarded: from the first transport the \textit{closing} action is erased, 
from all the intermediate actions both the \textit{opening} and \textit{closing} are removed,
and from the final \textit{transport} the \textit{opening} action is deleted. 
Figure \ref{fig:scenario2environmenttransform} shows the \textit{container-transform} being applied on an example task tree for fetching a spoon and a fork from a drawer.

The corresponding \textit{applicability} rule is shown below:

\begin{lstlisting}[style=Prolog]
container_rule(Open_path, Open_action, Close_path):-
  transport_task("top-level", First_task),
  transport_task("top-level", Second_task),
  ...,
  task_action_param(First_task, First_action),
  task_action_param(Second_task, Second_action),
  fetch_from_same_container(First_action,
                            Second_action),
  open_task(First_task, Open_task),
  task_action_param(Open_task, Open_action),
  task_path(Open_task, Open_path),
  close_task(First_task, Close_task),
  task_path(Close_task, Close_path).
\end{lstlisting}

The rule generates a triple of \textit{Open\_path}, \textit{Open\_action} and \textit{Close\_path}, which point to the subtasks of a single transporting task (\textit{First\_task}), but only if there is another existing transport that uses the same container. 
%
%
In contrast to the \textit{both-hands-transform}, the transformation function of \textit{container-transform} does not take the first set of bindings generated by the applicability rule, but a list of all possible bindings found by the predicate, i.e.\
a list of all transporting actions that share the same container is generated. 
If there are multiple containers and from each of them multiple transports take place, the function distinguishes between the different containers and handles them separately, such that each container is opened and closed exactly once.

\subsection{Transporting Objects With a Tray}

\begin{figure}[htb]
\vspace{-1ex}
\centering
\includegraphics[width=0.87\columnwidth]{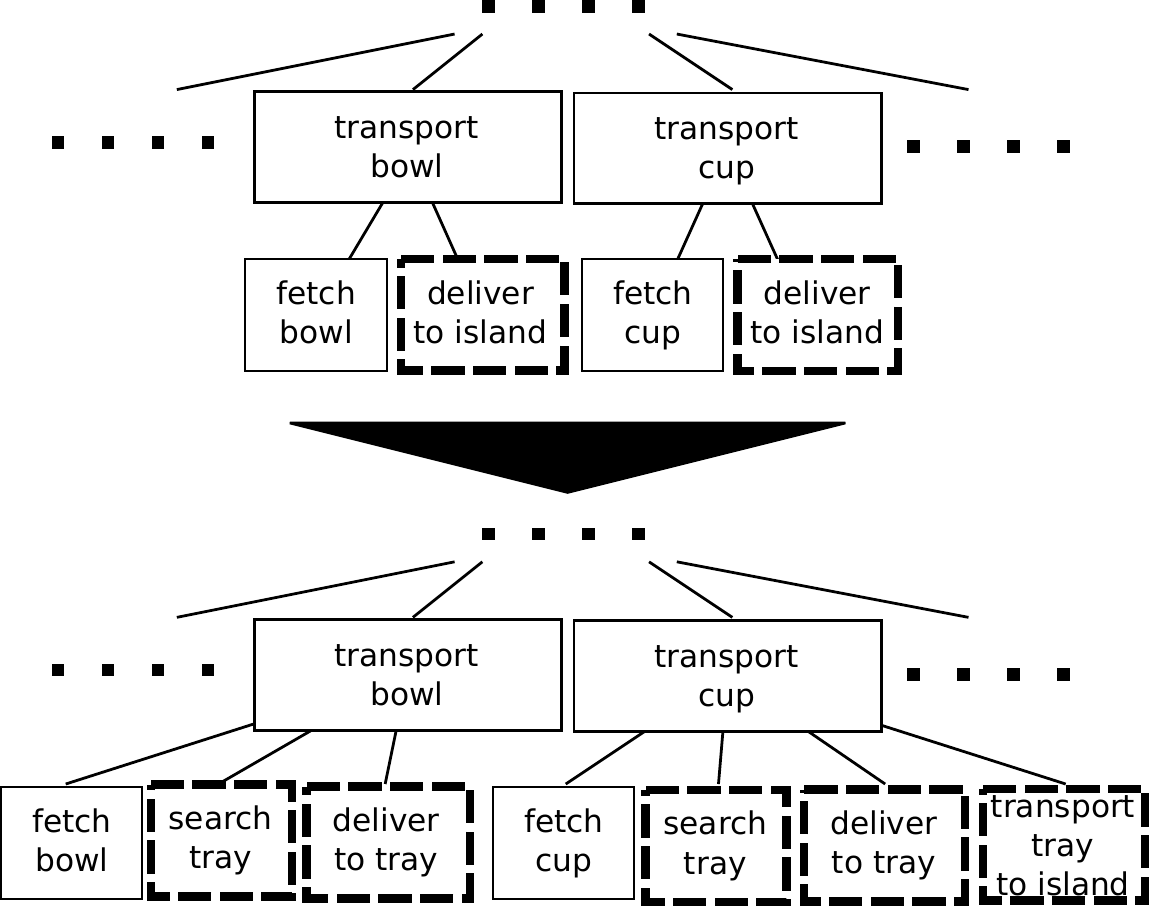}
\caption{Task tree for carrying objects from a certain area to the kitchen island transformed with \textit{tray-transform}. The delivery location of the objects is changed to a location on the tray, and a tray \textit{transporting} action is appended at the end of the last object transport.}
\label{fig:scenario1traytransform}
\vspace{-1.5ex}
\end{figure}

Whereas \textit{both-hands-transform} optimizes the plan to carry two objects at once, more than two objects can be transported by using a tray. 
Figure \ref{fig:scenario1traytransform} illustrates \textit{tray-transform} on an example of a task tree for transporting a bowl and a cup. 

The \textit{applicability} rule of the \textit{tray-transform} is a simplified version of \textit{both\_hands\_transform}, since both transformations require at least two transporting actions, that fetch and deliver objects to similar locations correspondingly. As the \textit{tray-transform} does not consider \textit{fetching} actions and only requires the paths and action descriptions of the \textit{deliver} tasks, the applicability predicate gets simplified.
The \textit{tray-transform} redirects the delivery locations of all applicable transporting actions to a location on the tray. To calculate this new location, the robot first searches the area for a tray and then places the object in a free spot on the tray. After all objects are delivered to the tray, the tray itself must be transported to the target location. Additional \textit{searching} and \textit{transporting} actions cannot be attached as new nodes to the task tree, since the hierarchy of the tree is restricted to the original plan's hierarchy. Instead, the actions of searching for the tray are appended at the front of the delivering actions, and the tray transport action is appended at the end of the \textit{code-replacement} of the last object delivering action.

%
%

\subsection{General Transformation Pipeline}
\label{sec:framework}

Transformations are applied by the robot to its plans automatically, which is implemented within the general transformation framework. Table \ref{tab:transformframework} lists its main API functions.

\begin{table}[htb]
    \vspace{-1.0ex}
    \centering
\begin{tabularx}{0.98\columnwidth}{|X|}
    \hline
    \multicolumn{1}{|c|}{\textbf{transformation-framework}} \\ \hline\hline
    register-transformation (transformation-function, applicability-rule) \\ \hline
    disable-transformation (transformation-function) \\ \hline
    enable-transformation (transformation-function) \\ \hline
    prioritize-transformation (superior-transform, inferior-transform) \\ \hline
    apply-all-transformations () \\ \hline
\end{tabularx}
\caption{API of the General Transformation Framework}
\label{tab:transformframework}
\vspace{-2ex}
\end{table}


The framework allows programmers to easily register new rules, which, once known to the system, will be used in the automatic transformation pipeline by the robot. Additionally, it allows to prioritize and organize the rules.
All suitable transformations are applied by calling the \textit{apply-all-transformations} function, which goes over the list of all registered transformations in a loop and applies them in the order of their defined precedence to the top-level task tree.

\section{EXPERIMENTAL EVALUATION}
\label{sec:experiments}

In this section, we evaluate our approach on a large amount of experiments in a fast plan projection environment to make statistically meaningful conclusions on improved execution performance, and demonstrate feasibility on a real-world table setting and cleaning plan performed by a PR2 robot.

\subsection{Experiments in Fast Plan Projection Environment}

\begin{figure}[htb]
  \centering
  \includegraphics[width=0.45\columnwidth]{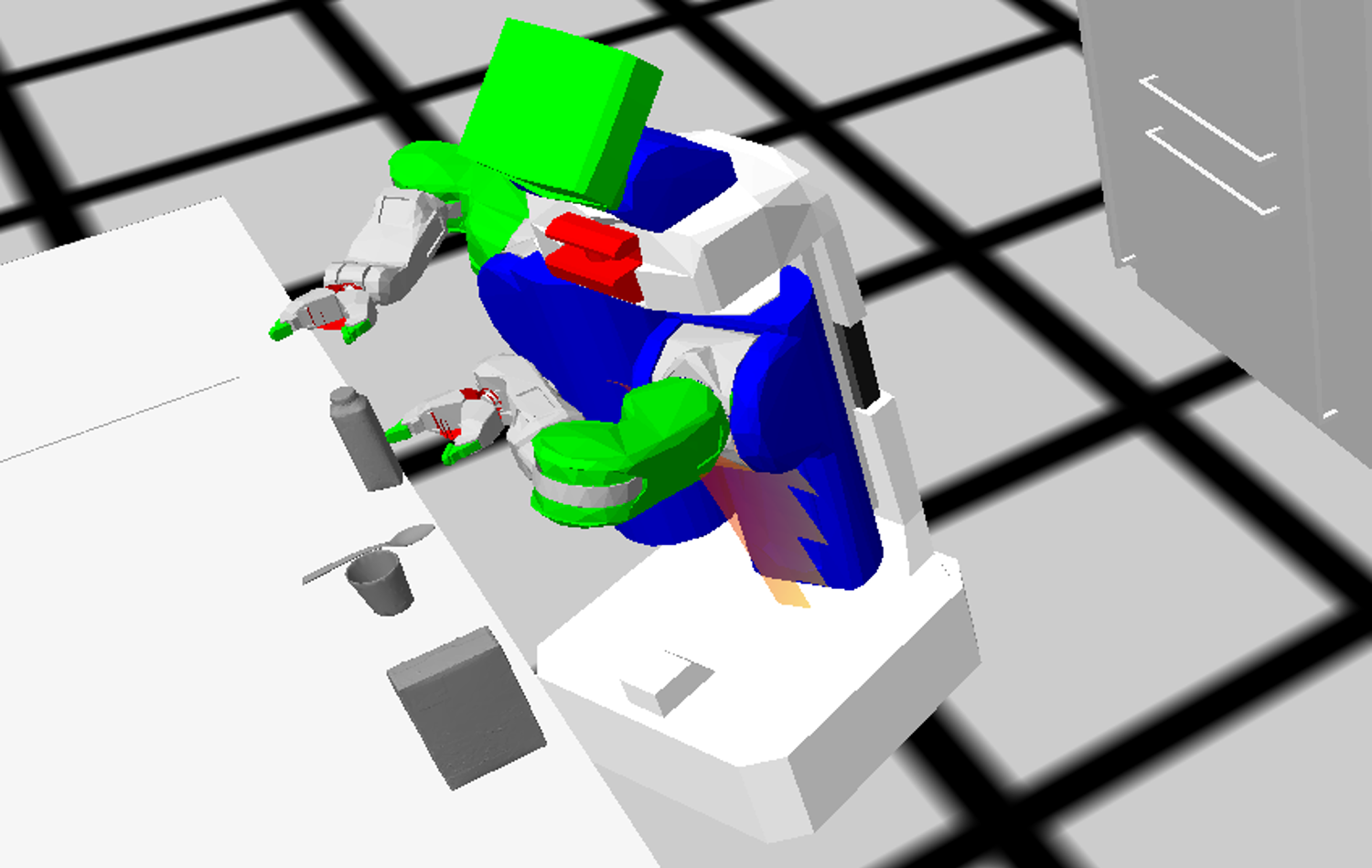} 
  \includegraphics[width=0.45\columnwidth]{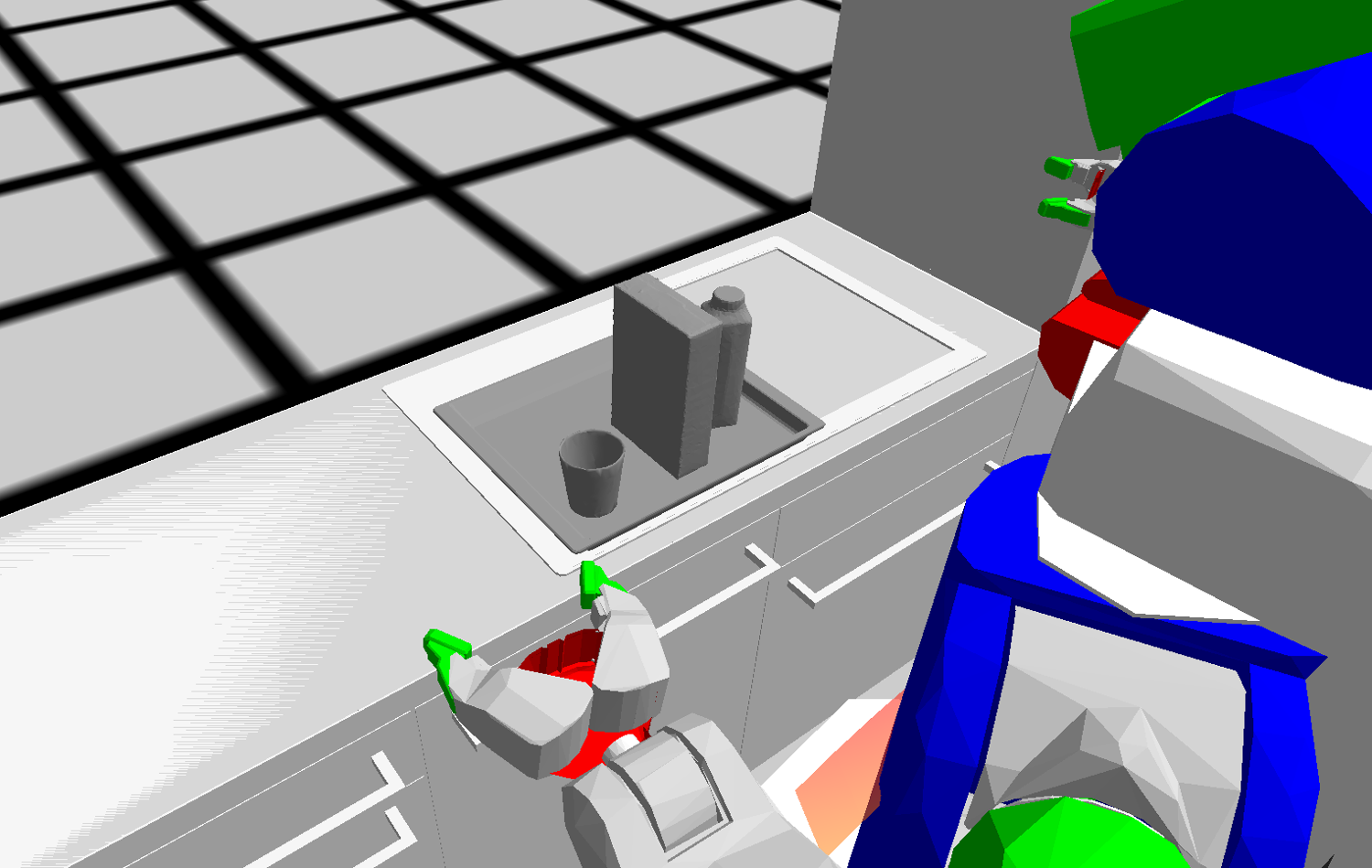}
  \caption{PR2 transporting objects in projection environment: (left) transporting four objects from kitchen island to the sink area with both hands, 
  (right) transporting three objects from the island to the sink with a tray}
  \label{fig:projection-demo}
  \vspace{-2ex}
\end{figure}

We evaluated \textit{both-hands-transform} on a scenario where four objects are transported from the kitchen island to the sink area (see Figure \ref{fig:projection-demo} (left)). The plan was executed 200 times with and without transformation each, with fixed initial and goal locations of the objects. The variety in execution originates from random sampling of grasps, arm to use, robot's base pose for fetching, searching and delivering, randomness factors of the IK solver, etc. We estimated the costs for navigation and manipulation by calculating euclidean distances between the key points of robot's base and gripper trajectories correspondingly. The number of performed actions and occurred failures has also been considered.


\begin{figure}[htb]
\centering
\includegraphics[width=0.98\columnwidth]{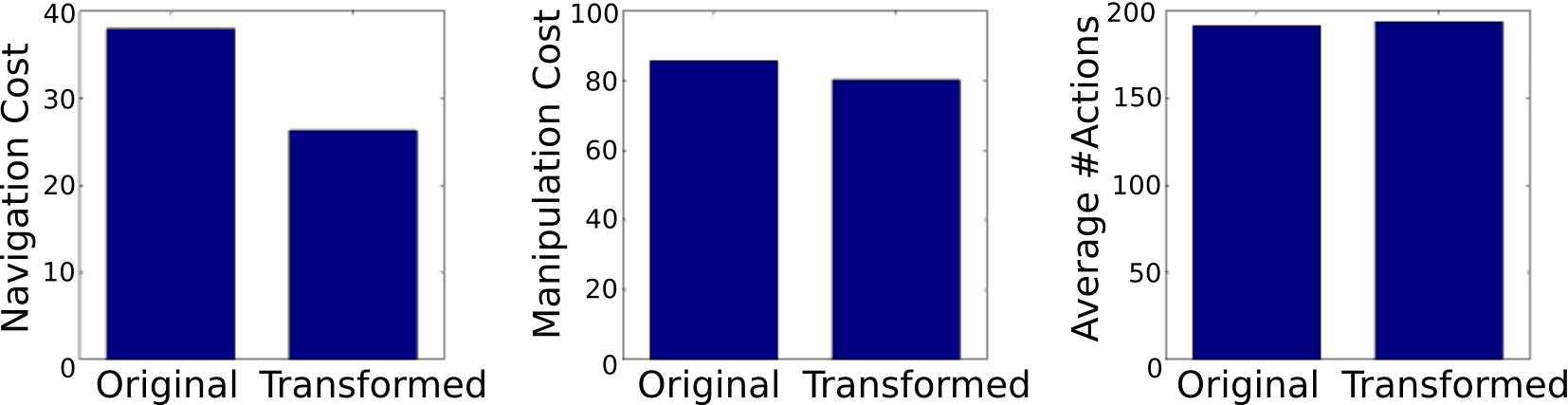}
\caption[Histogram scenario 1 both-hands-rule]{Evaluation data of a scenario for transporting four objects, where \textit{both-hands-transform} has been applied twice.}
\label{fig:hist4bh}
\end{figure}

With \textit{both-hands-transform} applied twice on a plan for transporting four objects, navigation distance decreased by 30\%, while manipulation cost stayed almost the same (see Figure \ref{fig:hist4bh}). The performance of the transformed plan improved with a significance value below 0.001, evaluated in a standard T-test. The transformed plan was on average 10.5\% prone to failure, as opposed to the original 6\%, since manipulating two objects at once introduces additional restrictions.

\begin{figure}[htb]
\centering
\includegraphics[width=0.98\columnwidth]{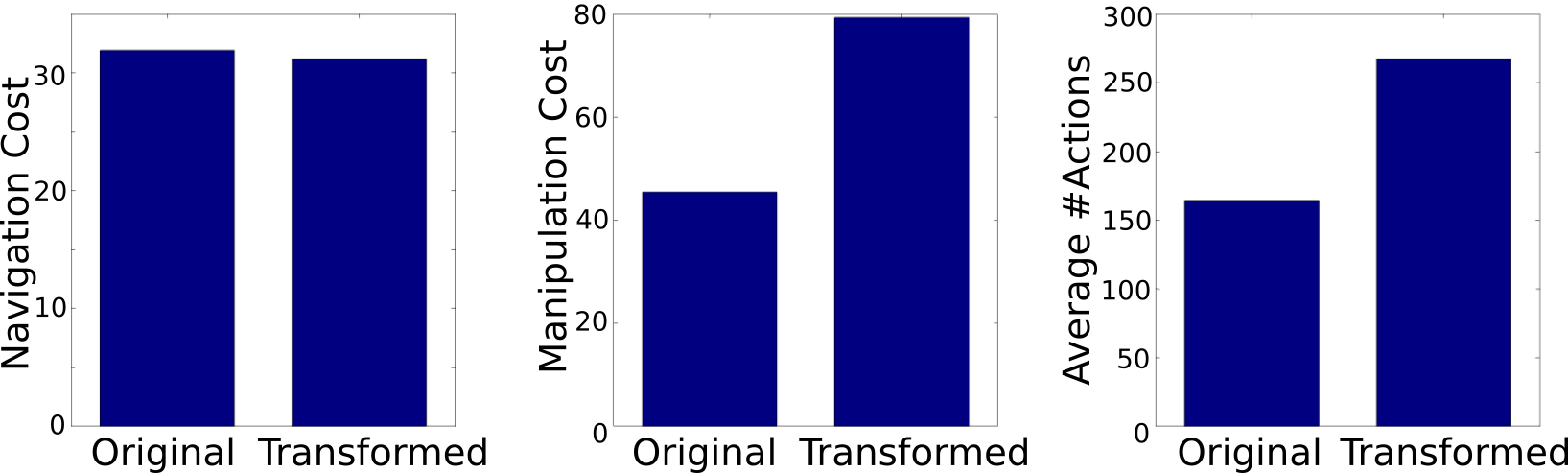}
\caption{Evaluation data of a scenario for transporting three objects, transformed with \textit{tray-transform}.}
\label{fig:hist3tray} 
  \vspace{-2ex}
\end{figure}

The \textit{tray-transform} was applied on a scenario with three objects (see Figure \ref{fig:projection-demo} (right)). Introducing complex additional actions, such as dual-arm manipulation of a tray, resulted in an increase in 
the manipulation cost, and the total amount of actions increased by 75\%, while the total navigated distance decreases only by 3\%  (see Figure \ref{fig:hist3tray}).
In our scenario, the starting and goal locations of transports were very close to each other, such that the navigation distance saved by using a tray could not outweigh the costs of additional actions.

The \textit{container-transform} has not been considered for empirical evaluation,
since removing a few actions from a plan obviously shortens the scenario. In projection, the robot does not move continuously but ``teleports'' through its via points, therefore, investigating the influence of leaving a drawer open on the navigation paths is not possible in projection. 

\subsection{Experiments on the Real Robot}

To show the feasibility of our approach, we designed a plan for the PR2 robot, where all three transformation rules can be applied.
The robot is supposed to transport four objects from one side of the kitchen to the other and back.
When executed with the general plan by a robot that is not equipped with a plan transformation framework,
first, two bowls are transported from the sink area surface to the kitchen island, one at a time. Then the robot fetches two spoons from the drawer and places them right of the bowls, whereby the drawer is opened and closed twice. When all four objects are set on the table, the robot carries them back to the sink area, one at a time.

Our PR2, equipped with the transformation framework, instead of executing the original plan right away, first runs the plan in projection and generates a task tree from it:

\begin{lstlisting}[style=Lisp]
(pr2-sim:with-projected-robot
  (set-table-and-clean-up))
\end{lstlisting}

Next, the robot transforms the generated task tree through a call to the general transformation framework:

\begin{lstlisting}[style=Lisp]
(transform-framework:apply-all-transformations)
\end{lstlisting}

All three transformation rules are applied, whereby the \textit{tray-transform} has a higher priority than the \textit{container-transform}, which has a higher priority than \textit{both-hands-transform}. Tray transformation is only applied, if there are more than two applicable transporting actions available, such that three or more objects are carried with a tray but two objects are simply carried in the two hands.

\begin{figure}[htb]
  \centering
  \includegraphics[width=0.32\columnwidth]{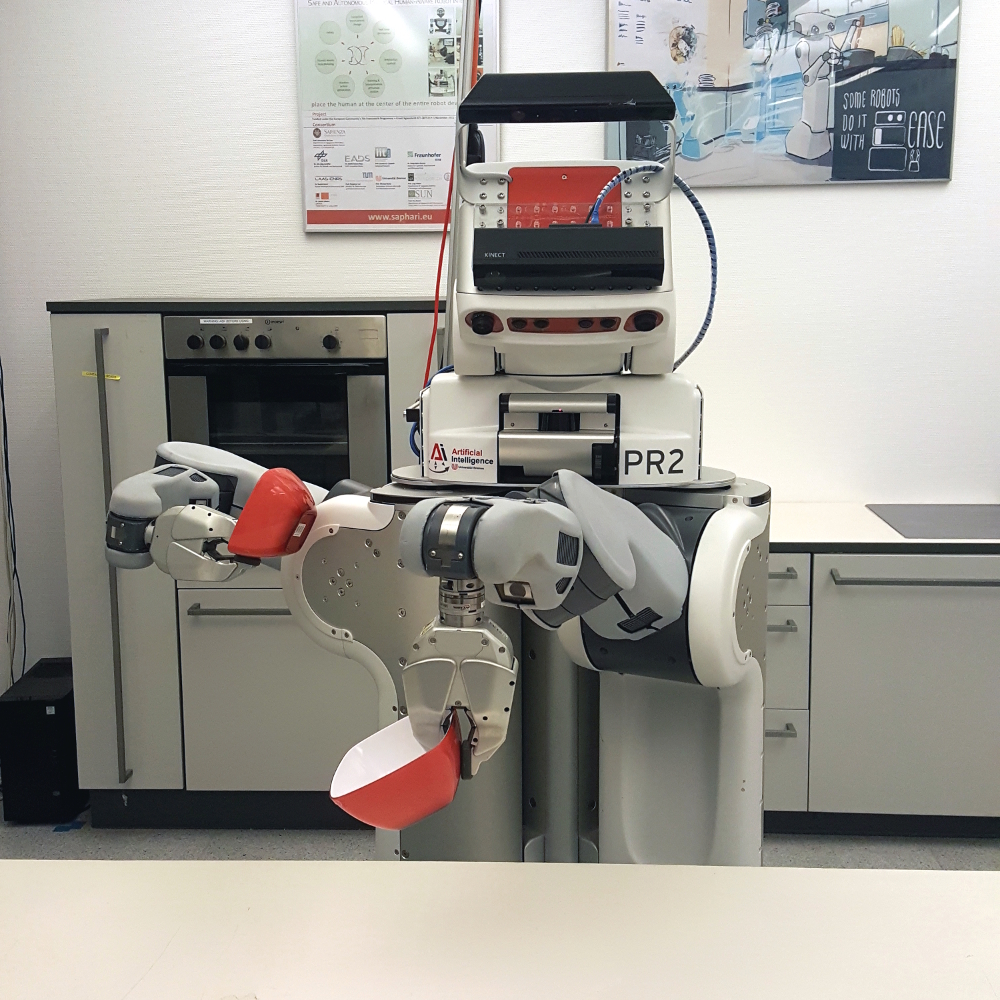} 
  \includegraphics[width=0.32\columnwidth]{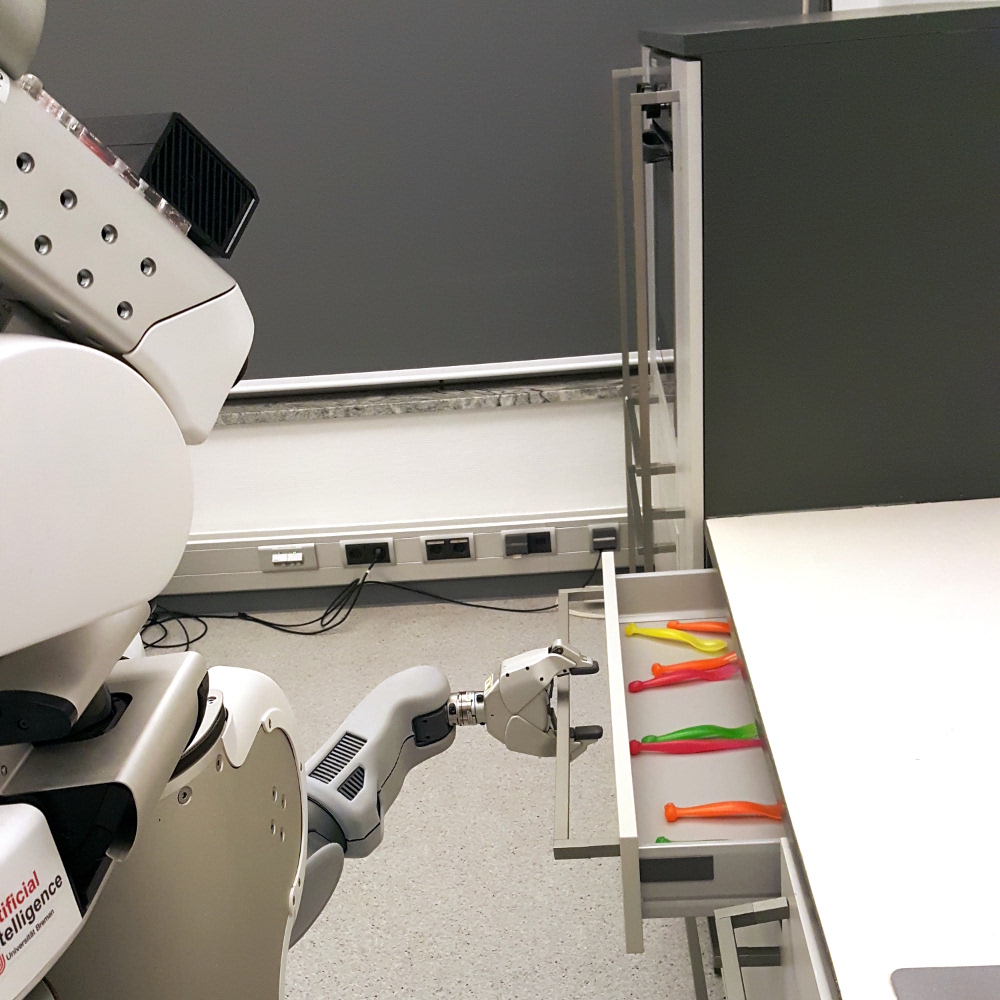}
  \includegraphics[width=0.32\columnwidth]{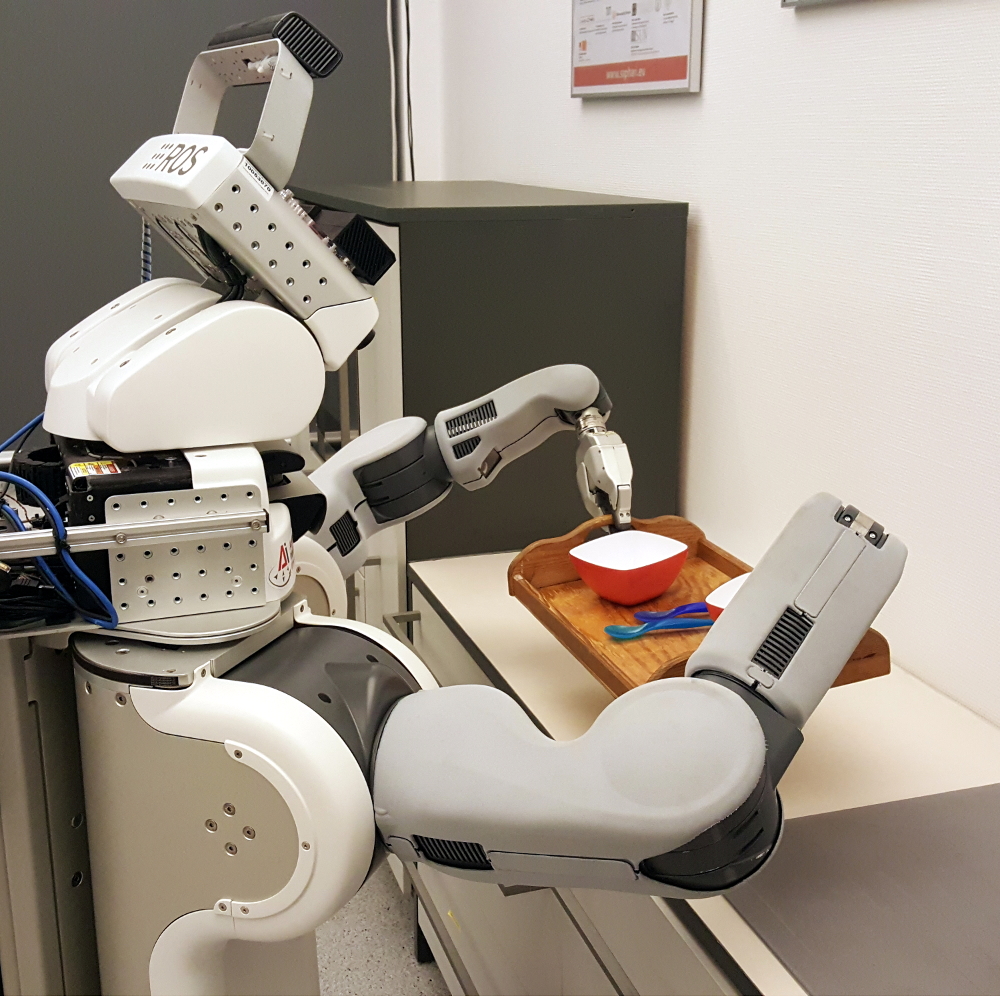}
  \caption{PR2 setting a table and cleaning up: (left) using both hands to transport two bowls, (middle) closing a drawer,
           (right) placing the tray.}
  \label{fig:real-robot-demo}
  \vspace{-2ex}
\end{figure}

Finally, the robot executes the same plan in the real world environment, and as the plan in projection is exactly the same as in the real world, the task tree is reused and the transformed version of the plan is executed:

\begin{lstlisting}[style=Lisp]
(pr2-interfaces:with-real-robot
  (set-table-and-clean-up))
\end{lstlisting}

As a result, PR2 uses both hands to bring the two bowls to the kitchen island at once (Figure \ref{fig:real-robot-demo} (left)), then it opens the drawer and closes it only after the second spoon has been transported (Figure \ref{fig:real-robot-demo} (middle)). To clean the table all four objects on the kitchen island are put on a tray, which is then carried to the sink area (Figure \ref{fig:real-robot-demo} (right)).

\section{RELATED WORK}

Multiple research groups investigated the possibilities of transformational planning. 
In the area of autonomous mobile manipulation,
TRANER ({TRA}nsformational Plan{NER} for Everyday Activity), introduced by M\"uller \cite{muller2008transformational}, is amongst the broader approaches on transformational planning towards offline plan refinement. Using RPL (Reactive Plan Language) \cite{rpl} for plan construction and transformation, the system operates on flexible plans for executing everyday tasks such as household chores. The system is evaluated in the Gazebo simulator \cite{gazebo}, where some physical limitations of robots and the environment are relaxed, e.g., the robot can extend its limbs arbitrarily, stacked objects are attached rigidly, etc., which allows for simpler plans.
From the area of automatic program transformations, Sussman's HACKER \cite{sussman} is a system which can change its programs when encountering a bug, and the knowledge from discovering the bug can be generalized and stored for future use. 
Transformations in HACKER result in programs with equivalent semantics. In robotics applications, transformations are applied to behavior specifications that have to be executable in perception-action loops, in which noisy sensor data and partial observability affect program semantics. 
%

Transformational planners such as Hammond's CHEF \cite{hammond} and Simmons's GORDIUS \cite{simmons} repair failing plans by searching for causal explanations of why a failure happened and replacing invalid assumptions in the plan.

In multi-robot systems, Bothelho and Alami \cite{Bothelho} present an approach for an architecture, where autonomous robots can cooperatively enhance their execution performance by allowing them to detect and recover from failures, focusing on resource conflicts between the robots.
%
Gateau et al.\ implemented a distributed architecture \textit{HiDDeN} \cite{Gateau}, which supports plan repair operations for failures concerning more than one robot. The repair strategies are kept as local as possible to avoid unnecessary multi-robot communications.

\section{CONCLUSION}
\label{sec:conclusion}


In this paper, we investigated transformation of plans, written with \textit{entity descriptions}.
We have shown that transformation techniques can be realized on plans designed for robots acting in the real world by using code replacements.
For utilizing the system on the robot, a plan is first projected, then automatically transformed and, finally, executed right away in the real world to produce optimized behavior.


Our approach with code replacements can be seen as ``patching'' the plan towards the situation at hand, and the robot does not store the transformed task tree. An interesting opportunity could be to use plan transformation techniques for long-term plan revision, where transformations could be applied to a complete category of tasks and the resulting plans could be stored for future reuse.

One limitation of our approach with code replacements is that the transformations cannot change the original structure of the plan. They can replace the code in single nodes of the task tree, but adding, removing or merging multiple nodes is not supported. Similarly, it is difficult to alter an already transformed task node, as that requires examining the code replacement of the node, which cannot be done robustly without considering all other code replacements in the tree. This limitation is due to the task tree being rigidly connected to the plan that generates it. As a result, the hierarchy of the tree cannot be altered. On the other hand, this also provides an advantage that we exploit greatly, which is the possibility to execute the same plan in different environments, while preserving the task tree and, therefore, its transformations, between the runs.
Another approach to plan transformation would be to alter the plan itself, i.e.\ the code directly, and not its runtime representation. This is a more challenging task that we will investigate in future work.


We have realized three example transformations: \textit{both-hands-transform}, \textit{container-transform} and \textit{tray-transform}.
Of course, there are a lot more possibilities that one can consider and we have shown just a few.
Discovering new opportunities for plan transformations is a promising direction for future work.
The three transformations that we have developed can be further improved in the future as well. For example, the current implementation of the \textit{container-transform} does not consider the possibility that leaving doors and cupboards open might block the robot's navigation paths or constrain its manipulation space.
Determining when to close a container and when not could be a complex issue that requires reasoning about navigation and manipulation trajectories.

The main goal of this paper was to investigate the possibility of applying plan transformations to general manipulation plans employed by robots acting in real-world environments, and we consider the paper to be an important step towards realizing robotic agents that can reprogram themselves despite their control structures being very complex due to the specifics of acting in the real world.

\section*{ACKNOWLEDGEMENTS}
\begin{small}
\noindent
This work was supported by DFG Collaborative Research Center \emph{Everyday Activity Science and Engineering (EASE)} (CRC \#1320).
\end{small}

\bibliography{bibliography}
\bibliographystyle{IEEEtran}

\end{document}